\documentclass{article}
\usepackage{amsmath}
\usepackage{amssymb}
\usepackage[final]{corl_2018}
\usepackage{graphicx}
\usepackage{booktabs} 
\usepackage{colortbl} 
\usepackage{xcolor} 
\usepackage{xfrac}
\usepackage{capt-of}
\usepackage{booktabs}
\usepackage{floatrow}
\usepackage{wrapfig}
\usepackage{todonotes}

\title{Sim-to-Real Reinforcement Learning for\\Deformable Object Manipulation}

\author{
  Jan ~Matas\\
  Department of Computing\\
  Imperial College London\\
  \texttt{jm6214@imperial.ac.uk} \\
  \And
  Stephen James \\
  Department of Computing \\
  Imperial College London \\
  \texttt{slj12@imperial.ac.uk} \\
  \And
  Andrew J. Davison \\
  Department of Computing \\
  Imperial College London \\
  \texttt{a.davison@imperial.ac.uk} \\
}
\begin{document}
\maketitle


\begin{abstract}
We have seen much recent progress in rigid object manipulation, but interaction with deformable objects has notably lagged behind. Due to the large configuration space of deformable objects, solutions using traditional modelling approaches require significant engineering work. Perhaps then, bypassing the need for explicit modelling and instead learning the control in an end-to-end manner serves as a better approach? Despite the growing interest in the use of end-to-end robot learning approaches, only a small amount of work has focused on their applicability to deformable object manipulation. Moreover, due to the large amount of data needed to learn these end-to-end solutions, an emerging trend is to learn control policies in simulation and then transfer them over to the real world. To-date, no work has explored whether it is possible to learn and transfer deformable object policies. We believe that if sim-to-real methods are to be employed further, then it should be possible to learn to interact with a wide variety of objects, and not only rigid objects. In this work, we use a combination of state-of-the-art deep reinforcement learning algorithms to solve the problem of manipulating deformable objects (specifically cloth). We evaluate our approach on three tasks --- folding a towel up to a mark, folding a face towel diagonally, and draping a piece of cloth over a hanger. Our agents are fully trained in simulation with domain randomisation, and then successfully deployed in the real world without having seen any real deformable objects.
\end{abstract}


\section{Introduction}
The majority of state-of-the-art work in robotic manipulation focuses on working with rigid objects, that either do not deform when they are grasped or have negligible deformation. However, deformable object manipulation has many important real-world applications. Key domains of interest are home assistance robotics (cloth folding~\citep{Miller2014AFolding}, bed making~\citep{Laskey2017LearningDART}, getting dressed~\citep{Gao2016IterativeInformation,TameiReinforcementRobot}); medicine (robot surgery~\citep{Thananjeyan2017MultilateralTensioning}, suturing~\citep{Schulman2013GeneralizationRegistration}); and industry (cable insertion~\citep{Vecerik2017LeveragingRewards}). Robots attempting to work with these objects are however presented with many new challenges, most notably the large object configuration spaces, the difficulty of accurate object behaviour modelling, and the large change in the configuration resulting from manipulation attempts.

Of the limited amount of work in deformable object manipulation, the majority focuses on folding 2D deformable objects, such as towels or articles of clothing. One approach employed explicit modelling of cloth deformation in simulation and then attempted to find an optimal trajectory based on the model~\citep{Li2015FoldingOptimization,Cusumano-Towner2011BringingPerception,yamakawaDynamicFold}. However, those models tend to be very sensitive to the deformation parameters of the objects (stiffness, shear resistance, friction) and therefore do not generalise well to unseen objects or environments. The second approach does not attempt to model the cloth but instead relies on visuomotor servoing to achieve the task. The robot identifies ideal grasping points based on heuristics (e.g. large curvature corresponds to a corner) and then executes a folding routine~\citep{Maitin-Shepard2010ClothFolding,osawaUnfolding,Bersch2011BimanualFolding}. Both approaches require a significant amount of engineering specific to the manipulation task, and it would be cumbersome to extend them to achieve success in a wholly different scenario. An alternative direction is to learn deformable object manipulation in an end-to-end manner, mapping observations directly to actions, and bypassing the need for explicit modelling. Specifically, we employ Reinforcement Learning (RL) to create an algorithm that is task agnostic and can learn many different behaviours based on the definition of a reward and a couple of provided demonstrations.  This has been extensively studied in the context of rigid object manipulation (see~\citep{Quillen2018DeepMethods} for a comprehensive evaluation), but only a small amount of work has focused on deformable objects. Moreover, no study has previously investigated the applicability of sim-to-real methods (such as domain randomisation) to transfer deformable object policies. We believe that if sim-to-real methods are to be employed further, then it should be possible to learn to interact with a wide variety of objects, and not only rigid objects, which has been the case to-date. To the best of our knowledge, deep RL and sim-to-real have not yet been applied to the domain of deformable object manipulation.

In this paper, we use an improved version of Deep Deterministic Policy Gradients (DDPG)~\citep{Lillicrap2015ContinuousLearning}, seeded with 20 demonstrations, to train an agent purely in simulation on three different tasks: folding a small towel diagonally, folding a towel up to a specific point and draping a towel over a small hanger. All tasks are learned via a single sparse reward on task completion. The agent receives only RGB images and the proprioceptive state (joint angles, gripper position) during test time. We employed domain randomisation~\citep{Tobin2017DomainWorld,jamesTransfer} in simulation to simplify the policy transfer from simulation to the real world without further training. Qualitative results can be seen in the video\footnote{\url{https://sites.google.com/view/sim-to-real-deformable}}.
\begin{figure}
    \centering
    \includegraphics[width=0.8\textwidth]{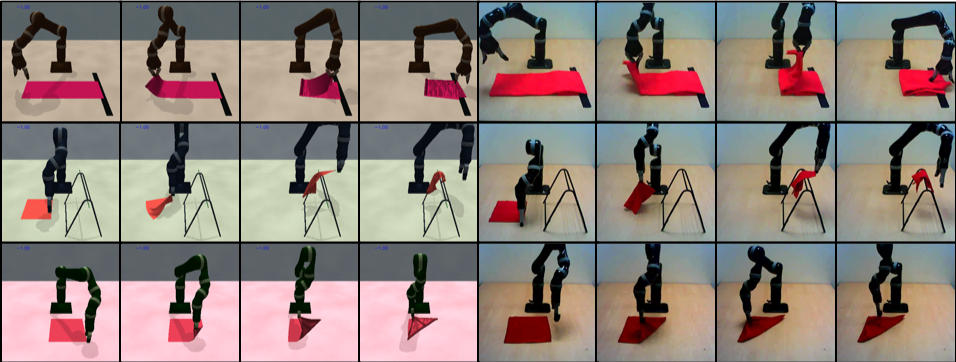}
    \caption{We learn robot policies in simulation and test them in the real-world. The algorithm was evaluated on 3 different tasks: folding a large towel up to a tape (top row), hanging a small towel on a hanger (middle row) and diagonally folding a square piece of cloth (bottom row).}
    \label{overwiev}
\end{figure}

\section{Related Work}
Cloth manipulation tasks solved by conventional robotics methods include cloth flattening~\citep{sunFlattening}, cloth folding~\citep{yamakawaDynamicFold,Li2015FoldingOptimization} or bringing cloth into a desired configuration~\citep{Cusumano-Towner2011BringingPerception}. The robots identify the cloth configuration based on visual information with hand-engineered heuristics and then use this either directly to parametrise a pre-programmed trajectory or indirectly by feeding the information to a mathematical model of the cloth. Some methods have also leveraged demonstrations for cloth manipulation, either through the use of behavioural cloning with noise injection~\citep{Laskey2017LearningDART, LeeLearningManipulation} or by creating a trajectory-aware registration method that becomes robust to distractions by observing the action multiple times~\citep{LeeLearningManipulation}. Other work has combined imitation learning and the PoWER RL algorithm to learn a policy for folding a towel by observing human demonstrations~\citep{balaugerFolding}. The towel was equipped with reflective markers and a complex system was employed to reconstruct the missing data if the markers were occluded or not detected.

Deep learning has not yet been extensively applied to cloth manipulation, even though it has found applications in many other robotic domains, including rigid object manipulation~\citep{guDDPG, Peters2008ReinforcementGradients}, UAV control~\citep{AbbeelAnFlight} or bipedal robot control~\citep{Peng2017DeepLoco:Learning}. One of the most prevalent deep RL methods in robotics is DDPG~\citep{Lillicrap2015ContinuousLearning}. The algorithm allows control in continuous space without discretisation, which makes it a good fit for controlling robot joint velocities. The algorithm has been the basis for a large number of extensions~\citep{Fujimoto2018AddressingMethods, AndrychowiczHindsightReplay, Barth-MaroDISTRIBUTEDGRADIENTS, Nair2017OvercomingDemonstrations, Schaul2015PrioritizedReplay, Pinto2017AsymmetricLearning} which have further improved the performance of the agent. DDPG can also be extended with demonstrations to considerably speed up  the learning process~\citep{Vecerik2017LeveragingRewards}.

Transferring policies learned in simulation into the real world is a challenging task. Previous work has shown that direct transfer was not possible~\citep{James20163DQ-Learning}, while others have shown that transfer only works after the agent has received additional training in the real world~\citep{rusuSimToReal}. A promising technique to accomplish a successful transfer from simulation to the real world is domain randomisation~\citep{Tobin2017DomainWorld, Pinto2017AsymmetricLearning,jamesTransfer}, which samples simulation parameters (e.g. camera position, light position, textures etc.) from probability distributions centred at a noisy estimate of the ground truth. As a result, the agent learns to ignore minor variations in the environment, so it becomes robust to domain changes, including the sim-to-real transfer. This approach has been successfully employed on a pick-and-place task, where an agent trained using supervised learning methods in the simulation was able to pick up a cube and move it to a basket, even in the presence of distractors and variable lighting~\citep{jamesTransfer}.
\begin{figure}
    \centering
    \includegraphics[width=1\textwidth]{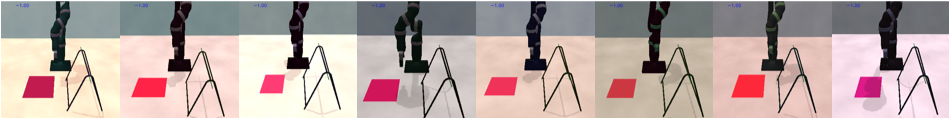}
    \caption{Examples of domain randomisation for the hanger environment. During randomisation, we vary the table textures, cloth and arm colours, light position, camera position and orientation, cloth size and position, hanger size and position, initial arm position and size of arm base.}
    \label{li_simulation}
\end{figure}
\section{Background}
\label{sec:background}
We consider a classic RL setting that can be represented as a Markov Decision Process (MDP) defined as a 5-tuple \((S, A, P, r, \gamma)\), where \(S\) is the set of full states of the environment, \(A\) is the set of actions,  \(r: S \times A \rightarrow \mathbb{R} \) is the reward function, \(\gamma\) is a discount factor and \(P: S \times A \times S : \mathbb{R} \) is a state transition probability function. The decision process is partially observable and the agent receives observations \(o\) from the set of observations \(O\). The reward function in our work is sparse, so the agent only receives the reward on accomplishing the full task. Hence, retraining the agent for a new task only requires defining the conditions of success and possibly a new hyper-parameter search, which can be parallelised and automated.

The goal of the agent is to learn a deterministic policy \(\pi : O \rightarrow A\) such that taking action \(a_t = \pi(o_t)\)  maximises the return from the state \(s_t\) (sum of discounted future rewards),  \(R_t = \sum_{i=t}^{\infty} \gamma^{i - t}r(s_i, a_i)\). After taking an action \(a\), the environment transitions from state \(s_t\) to state \(s_{t+1}\) sampled from probability distribution \(P(s_t, a_t, .) \). The quality of taking an action \(a_t\) in state \(s_t\) can be measured by a \(Q\) function \( Q(s_t, a_t) = \mathbb{E}[R_t| s_t, a_t]\).

DDPG~\citep{Lillicrap2015ContinuousLearning} is a deep RL algorithm for learning control policies in a continuous action domain. It uses an actor neural network, parametrised by a set of parameters \(\theta^\pi\), that maps observations to actions \(\pi: O \rightarrow A\) and tries to maximise \(Q (s_t, \pi(o_t)) \) at each time-step \(t\). However, the \(Q\) function is not known and DDPG employs a critic neural network, parametrised by parameters \(\theta^Q\), to estimate \(Q\) by minimising the Bellman loss: \[L_{critic} = (Q(s_t, a_t) - r_t - Q^\text{*}(s_{t+1}, \pi^\text{*}(o_{t+1}))) ^2 ~.\]

During training, the agent acts in the environment according to noisy policy \(a_t = \pi(o_t) + N(0, \sigma) \).  The Gaussian noise facilitates exploration. Each transition the agent generates is stored in a replay buffer from where it is sampled in batches to train the networks. Sampling from a replay buffer stabilises training by removing temporal correlations and therefore reduces the changes in the distributions the networks are trying to learn. DDPG also employs target networks \(Q^\text{*}\) and \(\pi^\text{*}\) to reduce the risk of Q-value estimates oscillating or diverging due to the recursive Q-value definition in the Bellman equation. 

DDPG became the primary building block of many other algorithms trying to improve on it. We give here a brief summary of the selected DDPG extensions that we incorporated into our algorithm.  
\paragraph{Prioritised Replay} Prioritised replay~\citep{Schaul2015PrioritizedReplay} assigns a priority \(p_i\) to each transition, computed as a sum of the last temporal difference (TD) error and small hyper-parameter \(\epsilon\). TD error is defined as the difference between critic prediction and critic target, so it serves as a proxy for the learning progress induced by the transition. \(\epsilon\) guarantees that even transitions with small TD errors can be sampled in the future, which is necessary because the critic changes its estimates as learning progresses. All new transitions are added to the replay buffer with priority equal to the current maximal priority in the buffer. The sampling probability is computed as \(P(i) = \frac{p_i^\alpha}{\sum_{k}{p_k^\alpha}}\), where \(\alpha\) is a parameter controlling the strength of the prioritisation. Prioritised sampling introduces a bias that needs to be corrected by multiplying the TD error of the transition when training by the importance sampling weight: \( w_i = (\frac{1}{NP(i)})^\beta\), where \(\beta\) is a hyper-parameter controlling the magnitude of bias correction and \(N\) is the replay buffer size.
\paragraph{N-Step returns} N-Step returns help to quickly propagate the reward signal throughout the robot trajectory by looking at N subsequent transitions instead of just one.  It has been shown to accelerate and stabilise learning~\citep{Barth-MaroDISTRIBUTEDGRADIENTS}. N-step returns change the critic loss to:
\[L_{nstep} = (Q(s_t, a_t) - \sum^N_{i=0}\gamma^ir_{t + i}  - \gamma^NQ^\text{*}(s_{t+N}, \pi^\text{*}(o_{t+N}))) ^2 ~.\]
It is possible to use both 1-step loss and N-step loss at the same time, in which case the critic loss becomes the sum of the losses weighted by two hyper-parameters \(\lambda_{nstep}\) and \(\lambda_{1step}\). 
\paragraph{DDPGfD} The original DDPG usually does not perform well on complex multi-step tasks with sparse rewards, because it is statistically improbable that the agent would often discover the right behaviour by random exploration. DDPGfD~\citep{Vecerik2017LeveragingRewards} overcomes this limitation by seeding the training with demonstrations, which are inserted into the prioritised replay buffer along with normal transitions. Demonstration transitions are never deleted from the replay buffer, and their priority is increased by a small constant \(\epsilon_D\) to make them more likely to be sampled. DDPGfD begins with a \textbf{pre-training phase}, where it executes a fixed number of training steps using the replay buffer initialised  with demo transitions. Following pre-training, it begins collecting new experiences. DDPGfD also employs N-Step returns and adds L2-regularisation on both actor and critic. 
\paragraph{Behavioural Cloning}~\citep{Nair2017OvercomingDemonstrations} 
DDPG can be further adapted to take advantage of demonstrations by introducing behavioural cloning loss to the actor network. This loss is applied only when a demonstration is sampled from the replay buffer for training. It encourages the actor to propose the same action as the demonstrator in the given state. After sufficient training, the agent might surpass the performance of the demonstrator and \(L_{BC}\) would then become detrimental to agent performance. The Q-filter mitigates this problem by only applying \(L_{BC}\) if the critic judges that the action proposed by the actor is worse than the action of the demonstrator. 
\[
    L_{BC}= 
\begin{cases}
    |\pi(o_i) - a_i|^2,& \text{if } Q(s_i, a_i) > Q(s_i, \pi(o_i))\\
    0              & \text{otherwise}
\end{cases}
\]
\paragraph{Reset to demonstration} Reset to demonstration~\citep{Nair2017OvercomingDemonstrations} aims to make it easier for the agent to receive a reward in sparse long-horizon tasks.  After the end of an episode, the environment will have a small probability of being placed into a random state encountered during demonstrations. In those cases, the agent only needs to complete the sub-task starting at the sampled state. This sub-task is usually substantially easier, particularly if the demonstration state was sampled near the end of the episode.
\paragraph{TD3} DDPG is prone to overestimating Q-values, which in turn leads to sub-optimal policies. TD3 ~\citep{Fujimoto2018AddressingMethods} implements 3 improvements to address the overestimation resulting from approximation errors. Firstly, it maintains 2 independent critic networks and always takes the minimum Q-value as the optimisation target for both actor and critic. Secondly, it proposes to delay the propagation of weight updates to target network by a couple of steps, so they have time to converge to a better quality update. Finally, it regularises the target Q-value by adding a clipped normal noise to the action proposed by the target actor to explicitly increase the smoothness of the Q-function prediction. The TD3 1-step target of the critic is defined to be:
\[y = r_t + min_{i=1,2}Q_i^\text{*}(s_{t+1}, \pi(o_{t+1}) + clip(\mathcal{N}(0, \sigma), -c, c))  ~.\]
\paragraph{Asymmetric actor-critic} The simulator always has a perfect understanding of the environment, which can be leveraged during the training phase. Asymmetric actor-critic~\citep{Pinto2017AsymmetricLearning} uses high dimensional (RGB) partial observations as an input to the actor, whilst using low-dimensional environment state (object positions, arm state, etc.) as the input for the critic. This extension significantly reduces the number of trainable parameters and increases the accuracy of the critic.

\vspace{-8px}
\section{Method}
\vspace{-5px}
\subsection{Simulation}
The reinforcement learning community currently uses many simulators to facilitate the cheap and fast collection of data. Among the widely used simulators, only Pybullet~\citep{Coumans2016PyBulletLearning} implements some rudimentary and experimental functionality for simulating deformable objects. Even though the simulator implements 2D rectangular cloth creation in its C++ API, we found the out-of-the-box simulation behaviour impractical for our purposes. We initially tried to rely on physics simulation to create a lasting grasp, which was not possible. The gripper either tunnelled through the cloth (low collision margin) or the gripper repelled it before the grasp attempt (high collision margin). We were only able to resolve the issue by creating a fake grasp implemented as a set of anchors between cloth nodes and gripper fingers.

The grasp creation was stochastic and deliberately failed in 5\% of the cases to expose the agent to unsuccessful grasp scenarios. Moreover, the creation of the constraint was subject to the gripper endpoint being in close proximity to a cloth node. Creating the constraint only to a single point on each gripper causes the cloth to spin unnaturally, so multiple anchors were used --- one at the middle and one at both extremities of each fingertip.  Finally, we found that the existing implementation of anchors between soft bodies and rigid bodies was not sufficient because it reached an equilibrium of forces with the cloth hanging approximately 5 cm below the gripper. We adapted the implementation so the anchor between cloth node and rigid object is honoured regardless of other forces acting on the cloth.

We employed domain randomisation to facilitate a smooth domain transfer of the learned policy. More specifically, we randomised the textures using Perlin noise ~\citep{Perlin1985AnSynthesizer}; object and background colours; object parameters and positions; arm spawn position and joint angles; camera position, orientation and intrinsics; light source position and colour; and all reflectance coefficients. The values were sampled from either normal or uniform distributions around the noisy ground truth estimates. 
\begin{figure}
    \centering
    \includegraphics[width=0.75\textwidth]{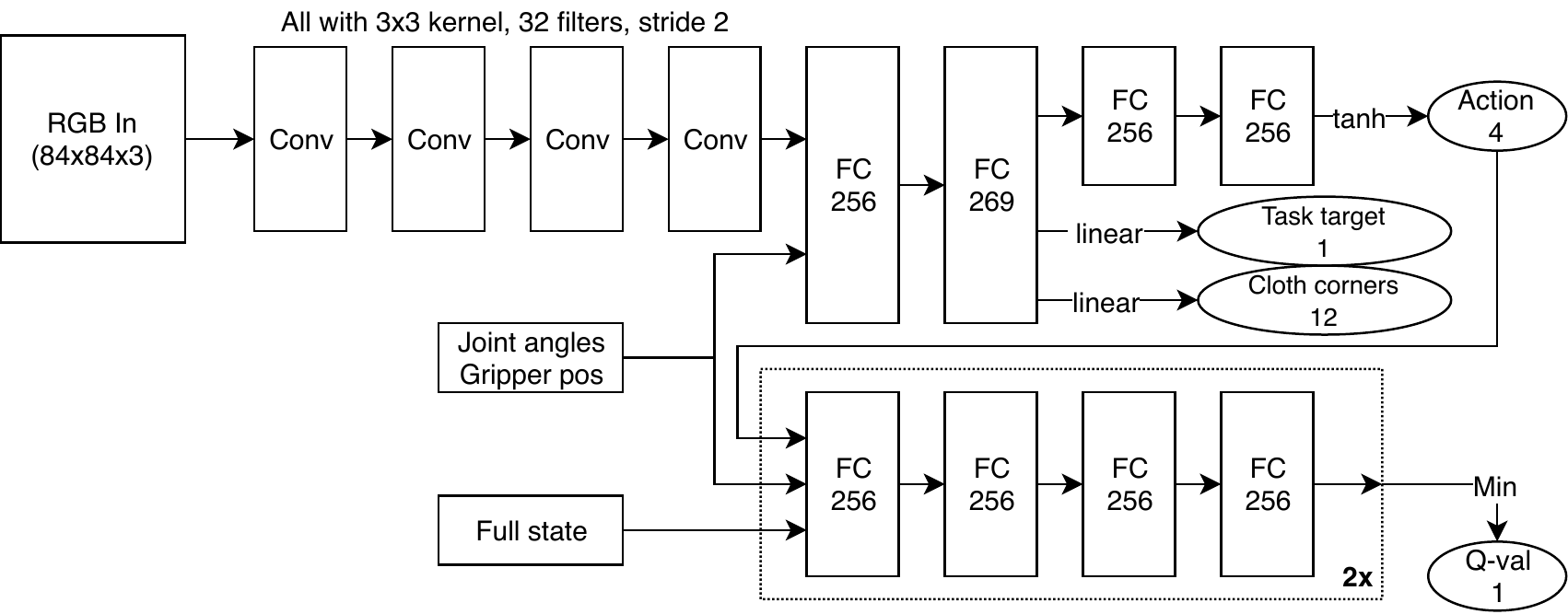}
    \caption{The network architecture uses 3 different inputs --- RGB images from the camera looking at the scene, joint angles and gripper position (available at test time from the robot API) and full state, which is only available at training time. The top half of the figure corresponds to the actor, while the bottom half corresponds to twin critics. The actor receives joint angles, gripper position and RGB images while the critic receives full-low dimensional state. Auxiliary outputs of the actor are only used during training to help the network quickly recognise essential scene features. However, they were also useful for debugging purposes at test time, because we can plot the estimate of cloth position and target position to verify that the actor understands the scene.}
    \label{network_arch}
\end{figure}

\subsection{Learning algorithm with integrated improvements}
During initial experimentation, we found that DDPG was not successful in solving any of the proposed environments, and so investigated possible improvements. We have taken inspiration from the success of the Rainbow DQN agent~\citep{rainbow} integrating all recently proposed extensions and achieving state-of-the-art performance on a set of benchmark tasks. Starting with the DDPG baseline available in the OpenAI repository~\citep{baselines}, we implemented all DDPG extensions listed in Section \ref{sec:background}. We however did not use the Q-value target regularisation in TD3 because we found it to be detrimental to the agent performance for these particular tasks. This results in the following critic loss, applied to \(a_t\) during training:
\begin{gather*}
L_{critic}(a) = \lambda_{nstep}L_{nstep}(a)w_i + \lambda_{1step}L_{1step}(a)w_i + \lambda_{L2}L_{reg}^Q(\theta^Q), \\
L_{nstep}(a) = (Q(s_t, a) - \sum^N_{i=0}\gamma^ir_{t + i}  - \gamma^Nmin_{i=1,2}Q_i^\text{*}(s_{t+N}, \pi^\text{*}(o_{t+N}))) ^2,  \\
L_{{1step}}(a) = (Q(s_t, a) - r_t - min_{i=1,2}Q_i^\text{*}(s_{t+1}, \pi^\text{*}(o_{t+1}))) ^2 
\end{gather*}
The auxiliary outputs predict the key features of the environments (in our case those are cloth corner positions, tape y-coordinate and hanger y-coordinate). \(L_{aux}\) is the mean square error between the prediction and the actual value. Each component of the auxiliary predictions can be weighted by separate weightings, although this was rarely used in practice. The resulting actor loss is:
\begin{gather*}
L_{actor} = -L_{critic}(\pi(o_t)) + \lambda_{BC}L_{BC} + L_{aux}\\
L_{BC} = 
\begin{cases}
    (\pi(o_i) - a_i)^2,& \text{if } Q(s_i, a_i) > Q(s_i, \pi(o_i)) \text{ and }i \text{ is demonstration}\\
    0              & \text{otherwise}
    ~.
\end{cases}
\end{gather*}

The priority of each transaction is updated after each training step according to:
\[p_i = L_{i,nstep}(a_i) + L_{i,1step}(a_i) + \epsilon + \epsilon_D\max_{k \in minibatch} (L_{k,nstep}(a_k) + L_{k,1step}(a_k))~,\]
where \(\epsilon = 10^{-6}\) is a small constant. We found that it was impossible to tune the fixed constant \(\epsilon_D\) (as suggested by DDPGfD) to boost the priority of demonstrations further because the TD error magnitude varied by multiple orders of magnitude across training epochs. We instead made the further demo priority boost term proportional to the maximal losses in the current mini-batch. \(\epsilon_D\) is set to 0 for updating priorities of all transitions apart from demonstrations. We used the same network architecture (Figure \ref{network_arch}) for all 3 experiments. The full learning algorithm with all improvements will be made available online.\footnote{\url{https://sites.google.com/view/sim-to-real-deformable}}.

\section{Experiments}
\subsection{Cloth manipulation environments}
All standard RL environments for manipulation tasks only contain rigid objects, so we designed and implemented 3 new environments for solving deformable object tasks. Each environment exposes an RGB observation with dimensions 84x84x3, a low dimensional state and low dimensional actor input (joint angles and gripper position). The robotic arm in the environments is 7DOF Kinova Mico controlled by 4-dimensional action. First 3 dimensions are the velocity of the end effector while the last dimension is a gripping velocity (negative for opening and positive for closing). The reward is sparse with +100 for success and 0 otherwise. Gripper rotation is not necessary for the tasks and is therefore kept fixed. The origin of the coordinate system is at the base of the arm, with z-axis perpendicular to the table and x-axis pointing towards the camera. The environments implement OpenAI gym~\citep{Brockman2016OpenAIGym} API and use Pybullet as a simulation engine~\citep{Coumans2016PyBulletLearning}. We call the 3 environments Tape, Hanging and Diagonal Folding:
\begin{enumerate}
\item \textbf{Tape}:  The robot needs to fold a large towel up to a mark identified by a piece of black tape. The tape can be in 3 different positions: 5/8th, 7/8th and at the end of the towel. The robot receives a reward if both corners of the lifted side of the cloth are within a threshold distance from the tape. The gripper is fixed to point downwards with fingers parallel to the y-axis. This task was proposed by \citet{LeeLearningManipulation}.
\item \textbf{Hanging}: The robot needs to grasp the piece of cloth and drape it over a small hanger. The cloth appears on the left side of the scene, and we sample its position from a uniform distribution. The reward is given when the cloth is released from the gripper, and all corners stay 5 or more cm over the ground for 20 simulation steps (this rules out cloth sliding off the hanger). The gripper has fingers parallel to the x-axis.
\item \textbf{Diagonal folding}: The robot needs to fold a rectangular face towel ($\sim 28 \times 28$cm) diagonally. The reward is given if the diagonal corners are within a threshold distance from each other and all pairs of corners on the same side of the rectangle are at distances larger than 3/4 of the side length when flat (this is to prevent the robot simply crumpling the cloth to align corners, which we have observed before). The gripper is parallel to the x-axis.
\end{enumerate}
\begin{figure}
    \centering
    \includegraphics[width=0.75\textwidth]{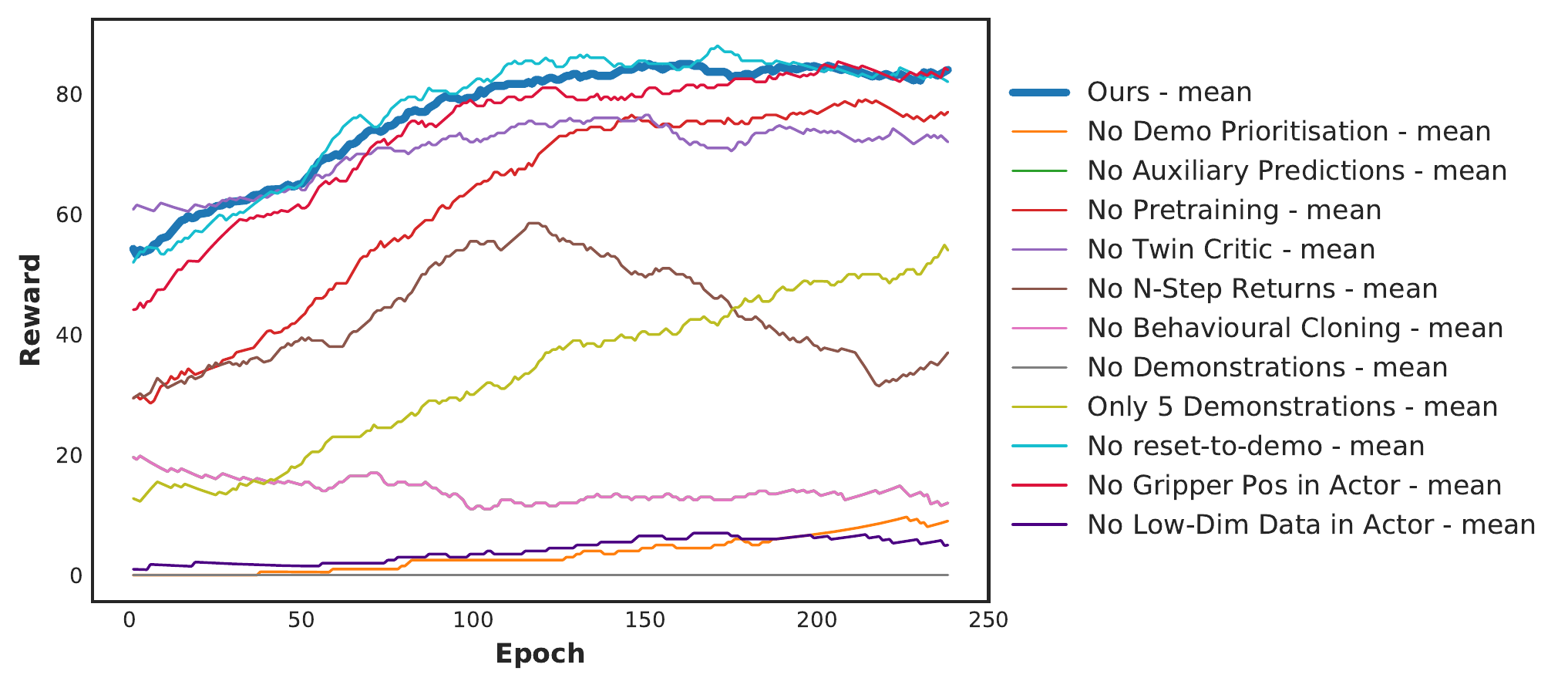}
    \caption{Ablation studies on the Diagonal Folding task, where ``Ours'' shows the result of the algorithm with all improvements. The reward for success was set to be 100, and therefore it is equal to the percentage of successes. Two evaluation episodes were performed after each epoch. Curves report the mean of 2 random seeds, and they were smoothed to improve legibility.}
    \label{ablations}
\end{figure}
\subsection{Simulation results and ablation studies}
 
\newlength{\oldintextsep}
\setlength{\oldintextsep}{\intextsep}

\setlength\intextsep{0pt}
\begin{wraptable}{r}{0pt}
\begin{tabular}{ccccc}
    \toprule
    \multicolumn{2}{c}{\textbf{Success rates (Sim)}} \\
    \midrule
    \rowcolor{black!20} Diagonal folding & $90\%$  \\
    Hanging & $77\%$  \\
    \rowcolor{black!20} Tape & $86\%$  \\
    \bottomrule
\end{tabular}
\caption{Success rates in simulation}
\label{sim-success}
\end{wraptable}
We ran the training algorithm with all implemented improvements (labelled ``Ours'') on the three tasks we defined above. Each training run was seeded with 20 demos. Each experiment took approximately 24 hours to run on one GeForce GTX TITAN.  The success rates (mean of 3 random seeds) in the final evaluations of the experiments are shown in Table \ref{sim-success}, which were achieved after approximately 80k transitions and in the presence of domain randomisation.

The most likely failure case across all environments is a failure to grasp the cloth. Even though the agent has learned to do multiple re-grasps, in some situations, it repeatedly fails (e.g. by closing the gripper above the towel). We believe this is due to an outlier in the camera configuration sampled from a normal distribution. Secondly, too fast or inaccurate motion usually causes the agent to crumple the towel after which it is no longer able to achieve the task. Thirdly, in the Hanging task, the agent sometimes drapes the cloth too far, causing it to fall. 

We performed ablation studies to verify the contributions of selected modifications to DDPG. The agent integrating all improvements either outperforms or matches the performance of all training runs with an ablation. Two implemented improvements do not seem to increase the agent performance:  reset to demonstration and adding gripper position to the low dimensional actor input. In the first case, we hypothesise that due to the BC loss, the agent can complete successful full-length tasks early in training so it can quickly form a diverse set of successful episodes. This might be preferable over repeatedly resetting to similar states from demonstrations. In the second case, we removed gripper position from actor input, instead making it an auxiliary output. The agent accurately learned to predict the forward kinematics, so the gripper position input was not necessary. However, also removing joint angles (No Low-Dim Data in actor) was detrimental to performance which indicates that the agent cannot infer gripper position accurately from images only.

The features with questionable value are Twin Critic and Pre-training. Although they seem to provide improvement, the trade-off is increased computational cost. Pre-training has a constant cost of 7 minutes at the start of training and maintaining two critics increased runtime by 1\%. However, Twin Critic would be substantially more expensive if it also used RGB observations. 

The improvements that convincingly demonstrated a positive contribution to agent performance are Auxiliary predictions, Behavioural Cloning and Demo prioritisation. Without boosting the priority of the demonstrations (adding the \(\epsilon_D\) term to priority equation), they are much less likely to be sampled because they form only a tiny portion of the replay buffer.

\subsection{Sim-to-real transfer}
\begin{table}

\begin{minipage}{.33\linewidth}
\centering
\medskip

\begin{tabular}{ccccc}
    \toprule
    \multicolumn{2}{c}{\textbf{Hanging task}} \\
    \midrule
    \rowcolor{black!20} Vicinity & $100\%$  \\
    Grasp & $76.6\%$  \\
    \rowcolor{black!20} Drape over & $70\%$  \\
    Full success & $46.6\%$  \\
    \bottomrule
\end{tabular}
\end{minipage}\hfill
\begin{minipage}{.33\linewidth}
\centering
\medskip

\begin{tabular}{ccccc}
    \toprule
    \multicolumn{2}{c}{\textbf{Diagonal folding task}} \\
    \midrule
    \rowcolor{black!20} Grasp & $66.6\%$  \\
    Not crumpled & $66.6\%$  \\
    \rowcolor{black!20} d $\leq$ 0.15m & $53.3\%$  \\
    d $\leq$ 0.1m & $40\%$  \\
    \rowcolor{black!20} d $\leq$ 0.05m & $20\%$  \\

    \bottomrule
\end{tabular}
\end{minipage}\hfill
\begin{minipage}{.33\linewidth}
\centering

\medskip

\begin{tabular}{ccccc}
    \toprule
    \multicolumn{2}{c}{\textbf{Tape folding task}} \\
    \midrule
    \rowcolor{black!20}Grasp & $90\%$  \\
    d $\leq$ 0.15m & $90\%$  \\
    \rowcolor{black!20} d $\leq$ 0.1m & $76.6\%$  \\
    d $\leq$ 0.05m& $43\%$  \\
    
    \bottomrule
\end{tabular}
\end{minipage}
\caption{The success rates for each environment in the real world. Note that these are run in the real world without additional training. For the hanging task, \textit{vicinity} means the gripper being within 5cm from the cloth, \textit{drape over} means the cloth is touching the top part of the hanger and \textit{full success} is achieved if the cloth does not fall after it is released. For diagonal folding, \textit{not crumpled} means that adjacent corners are more than 15cm from each other and the $d$ is the distance between diagonal corners (lower is better). For tape folding, $d$ is the distance between towel edge and the tape mark. }
\label{real-success}
\vspace{-5px}
\end{table}

In real-world experiments, we use the Kinova Mico 7DOF robotic arm mounted in the middle of a table, and we collect the RGB observation using a low-cost Genius C170 web camera mounted on a fixed tripod next to the table. We report the results of 30 trials on the real robot for each task in Table \ref{real-success}. As in simulation, the most prominent failure case is failed grasping, particularly with thin face towels (used in Hanging and Diagonal folding). The robot has only a small acceptable margin of error (roughly 1 cm) in the z-axis for a successful grasp --- going too low will prevent the gripper from closing and going too high will not grasp the cloth. The other common failure case was an imprecise movement resulting in crumpling of the fabric from which the agent was not able to recover. this was partly caused by low simulation fidelity. The real cloth was much stiffer and therefore less forgiving to imprecise movement and the agent could not learn this in simulation.

When experimenting with various levels of domain randomisation, we found that heavy randomisation can be detrimental to learning. Specifically, we tried sampling the texture colours from a uniform distribution across all colours and the performance of the agent after the transfer was significantly worse. We believe that it then became much harder for the network to identify invariant environment features it could use for orientation. Consistently with previous work ~\citep{jamesTransfer}, we found that camera randomisation is essential for successful transfer. Even with randomisation, the agent was still very sensitive to the camera position.

\section{Conclusion and Future work}
Building up on recent work in end-to-end learning for rigid object manipulation, we have extended those ideas to the domain of deformable objects and specifically, we have addressed the problem of cloth manipulation. We proposed a task agnostic algorithm based on Deep RL which bypasses the need to explicitly model cloth behaviour and does not require reward shaping to converge. The agent was able to learn 3 long horizon tasks: folding a towel to a tape mark, diagonal folding of face towel and draping a small towel over a hanger. Training was seeded with 20 demonstrations and happened entirely in simulation with a couple of adaptations to account for imperfections in experimental deformable body support, and with domain randomisation to enable easy transfer of the policy. The learning algorithm incorporated 9 improvements proposed in the recent literature and we have presented ablation studies to understand the role of these improvements. 

We believe that the primary factor limiting further research into deformable object manipulation is the lack of support for those objects in most robotic simulators. We are hoping that further research into simulation will allow us to create an accurate model of deformable object grasping, incorporate it into a widely used simulator and release the environments to create a set of benchmark tasks for future research in the domain.

\clearpage


\bibliography{corl}  

\end{document}